\documentclass[10pt,twocolumn,letterpaper]{article}

\usepackage[pagenumbers]{cvpr} 

\usepackage[dvipsnames]{xcolor}

\definecolor{cvprblue}{rgb}{0.21,0.49,0.74}
\usepackage[pagebackref,breaklinks,colorlinks,citecolor=cvprblue]{hyperref}

\def\paperID{5133} 
\def\confName{CVPR}
\def\confYear{2024}

\usepackage{hyperref}
\usepackage[utf8]{inputenc} 
\usepackage[T1]{fontenc}    
\usepackage{hyperref}       
\usepackage{url}            
\usepackage{booktabs}       
\usepackage{amsfonts}       
\usepackage{nicefrac}       
\usepackage{microtype}      
\usepackage{xcolor}         

\usepackage{tabularx}
\usepackage{array}
\usepackage{booktabs}
\usepackage{enumerate}
\usepackage{enumitem}
\usepackage{amssymb}
\usepackage{amsmath}
\usepackage{bbm}
\usepackage{graphicx}
\usepackage{xspace}

\makeatletter
\DeclareRobustCommand\onedot{\futurelet\@let@token\@onedot}
\def\@onedot{\ifx\@let@token.\else.\null\fi\xspace}
 
\def\eg{\emph{e.g}\onedot} 
\def\ie{\emph{i.e}\onedot}

\makeatother

\title{Generative 3D Part Assembly via Part-Whole-Hierarchy Message Passing}

\author{Bi'an Du$^1$, Xiang Gao$^1$, Wei Hu$^1$\thanks{Corresponding Author: Wei Hu (forhuwei@pku.edu.cn).}, Renjie Liao$^{2,3,4}$\\
$^1$Wangxuan Institute of Computer Technology, Peking University\\
$^2$University of British Columbia, $^3$Vector Institute for AI, $^4$Canada CIFAR AI Chair\\
{\tt\small pkudba@stu.pku.edu.cn, gyshgx868@pku.edu.cn, forhuwei@pku.edu.cn, rjliao@ece.ubc.ca}
}

\begin{document}
\maketitle

\pagestyle{empty}  
\thispagestyle{empty} 

\begin{figure*}
  \centering
  \includegraphics[width=1.0\textwidth]{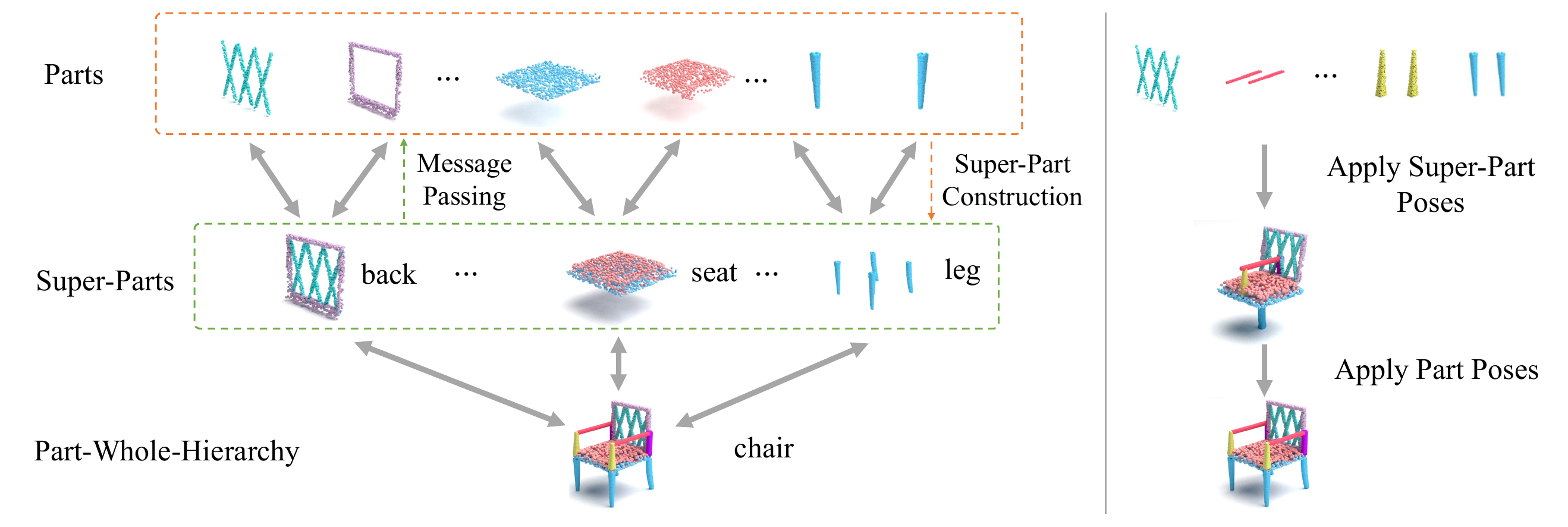}
  \vspace{-8mm}
  \caption{Left: an illustration of the part-whole-hierarchy for 3D shapes; Right: the part assembly process via the proposed part-whole-hierarchy message passing network.}
  \vspace{-3mm}
  \label{fig:teaser}
\end{figure*}

\begin{abstract}
Generative 3D part assembly involves understanding part relationships and predicting their 6-DoF poses for assembling a realistic 3D shape. 
Prior work often focus on the geometry of individual parts, neglecting part-whole hierarchies of objects. 
Leveraging two key observations: 1) super-part poses provide strong hints about part poses, and 2) predicting super-part poses is easier due to fewer super-parts, we propose a part-whole-hierarchy message passing network for efficient 3D part assembly.
We first introduce super-parts by grouping geometrically similar parts without any semantic labels. 
Then we employ a part-whole hierarchical encoder, wherein a super-part encoder predicts latent super-part poses based on input parts. 
Subsequently, we transform the point cloud using the latent poses, feeding it to the part encoder for aggregating super-part information and reasoning about part relationships to predict all part poses.
In training, only ground-truth part poses are required. 
During inference, the predicted latent poses of super-parts enhance interpretability. 
Experimental results on the PartNet dataset show that our method achieves state-of-the-art performance in part and connectivity accuracy and enables an interpretable hierarchical part assembly. Code is available at \href{https://github.com/pkudba/3DHPA}{https://github.com/pkudba/3DHPA}.
\end{abstract}    
\section{Introduction}
\label{sec:introduction}

Generative 3D Part Assembly \cite{litvak2019learning, zakka2020form2fit, shao2020learning, luo2019reinforcement, funkhouser2004modeling} is an emerging research area that aims to generate complex 3D shapes via assembling simple 3D parts without relying on prior semantic knowledge. 
Different from traditional 3D shape generation, it focuses on generating diverse, plausible configurations of given parts.
It facilitates the generation of complex objects and scenes with compositionality, flexibility, and efficiency. 
With the rapid evolution of 3D printing technology and the increasing demand for diverse 3D shapes, generative 3D part assembly finds applications in various scenarios, attracting attention from experts in computer vision, graphics, robotics, and machine learning.

However, achieving generative 3D part assembly presents a significant challenge due to the vast number of potential part arrangements and orientations, coupled with the intricate dependencies among the parts. 
Adding to the complexity, part geometry exhibits notable variation even within the same object category, making it exceedingly difficult to generalize learned assembly patterns across different objects.

Previous approaches primarily focus on designing architectures capable of learning powerful representations for individual 3D parts.
The hope is that these learned representations could facilitate accurate part assembly, either in a one-shot manner or sequentially \cite{li2020learning,zhan2020generative,narayan2022rgl,zhang20223d}. 
However, these methods often ignore the inherent part-whole hierarchies in 3D shapes in representation learning. 
For instance, as illustrated in Figure \ref{fig:teaser}, a chair consists of \emph{super-parts} such as seats, backs, and legs, with each super-part further divisible into \emph{parts} like seat surfaces and seat frames. Understanding the pose of a super-part provides insights into the poses of constituent parts within the same super-part, as they often share similar orientations or exhibit symmetry (\eg, left-right symmetry in chair legs and arms).
Moreover, predicting super-part poses is typically easier due to the fewer number of super-parts compared to parts.
Incorporating these hierarchies into the modeling would make the learning process and potentially improve the overall performance. 



In this paper, we introduce a part-whole-hierarchy message passing network for 3D shape assembly, predicting 6 degrees of freedom (6-DoF) poses for super-parts and parts in a hierarchical manner. 
We establish the correspondence between parts and super-parts (\ie, subsets of parts) by grouping parts based on their geometric similarities, following the approach from \cite{zhang20223d, zhan2020generative}.
Importantly, we treat super-part poses as latent variables to be learned, thereby eliminating the requirement of ground-truth super-part poses in our work. 
Our model comprises two sequential modules: the super-part encoder and the part encoder.

The first super-part encoder takes the 3D point clouds of all parts as input and employs the attention-based message passing to predict super-parts poses.
The representation of each super-part is aggregated from representations of parts within it.
These super-part poses provide initial estimation for the poses of individual parts.

Subsequently, the whole point clouds are transformed based on the predicted super-part poses and fed to the part encoder. 
This module, using a cross-attention mechanism, extracts features from the transformed point cloud, transferring super-part level information to the part level. 
It then leverages the attention-based message passing again to capture relationships among individual parts.

Following previous works~\cite{li2020learning,zhan2020generative,narayan2022rgl,zhang20223d}, we train and evaluate our model on the PartNet dataset \cite{mo2019partnet}.
Overall, our model achieves the state-of-the-art performances, outperforming the strongest competitor \cite{zhang20223d} by a significant margin, \ie, with an almost 2$\%$ improvement in mean part accuracy and a 3$\%$ improvement in mean connectivity accuracy. 
Moreover, through visual analysis of the assembly process for both super-parts and parts, we not only showcase accurate generation of part poses but also demonstrate interpretability via the predicted super-part poses. 
This interpretability feature further improves the utility of our model.


\section{Related Work}
\label{sec:related}

\begin{figure*}
  \centering
  \includegraphics[width=1.0\textwidth]{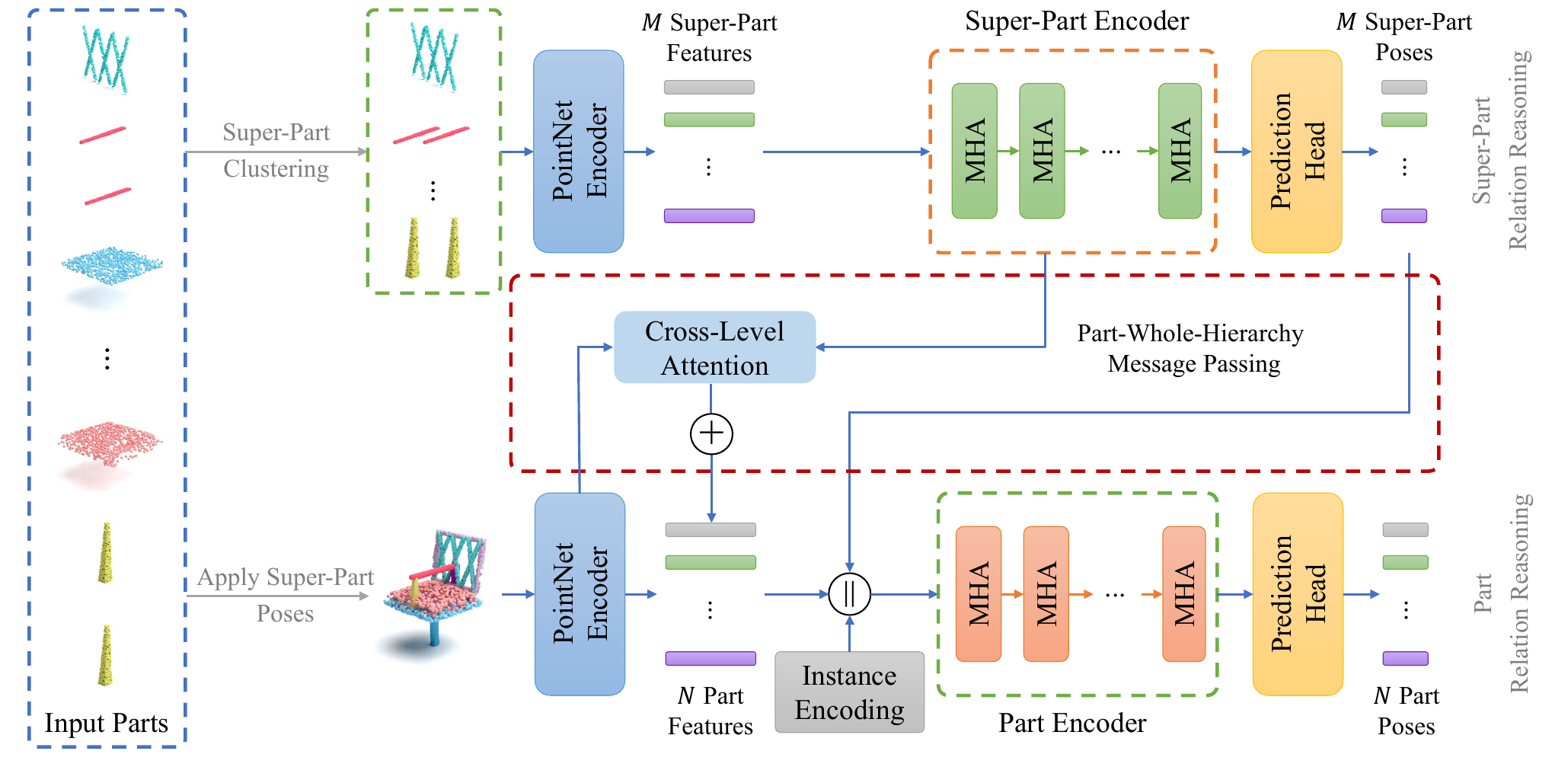}
  \vspace{-8mm}
  \caption{The overall architecture of our model consists of two modules: \emph{super-part encoder} and \emph{part encoder}. We first obtain super-parts via grouping parts based on their geometric similarities in an unsupervised fashion. The super-part encoder takes point cloud as input and predicts latent super-part poses (\textbf{no ground truth is needed}). The point cloud is then transformed based on super-part poses and fed to the part encoder. We incorporate both cross-level and within-level attention in the part encoder to predict part poses. }
  \vspace{-3mm}
  \label{fig:architecture}
\end{figure*}

\paragraph{Assembly3D modeling}

Numerous prior studies have approached the challenging 3D part assembly through the joint estimation of part poses.
The influential work by \cite{funkhouser2004modeling} introduces an intelligent scissoring technique to tackle this problem for part components.
Subsequent research by \cite{chaudhuri2011probabilistic, kalogerakis2012probabilistic, jaiswal2016assembly} employs graphical models to capture the semantic and geometric relationships among shape components, enabling exploration in assembly-based shape modeling.
PAGENet \cite{li2020learning} presents a network that is aware of individual parts and generates semantic parts along with their poses. 
\cite{xu2023unsupervised} proposes to decompose shapes using a library of 3D parts provided by the user.
However, these studies either assumed prior knowledge of part semantics or relied on existing shape databases. 
In a more practical setting, the authors in \cite{zhan2020generative, narayan2022rgl, zhang20223d, chen2022neural} focus on the pose estimation of individual parts without relying on shape databases or known semantic information. 
Specifically, DGL \cite{zhan2020generative} employs a dynamic part graph to iteratively refine the poses of individual parts. 
A progressive strategy using the recurrent graph learning framework has been investigated in \cite{narayan2022rgl}. 
Additionally, the authors in \cite{zhang20223d, chen2022neural} utilizes the Transformer \cite{vaswani2017attention} to model the structural relationships and performs the simultaneous assembly of all parts.
We follow this line of work and propose novel improvement to generate structurally-coherent part assemblies.


\paragraph{Structural Shape Generation}

In recent years, deep generative models, such as generative adversarial networks (GANs) \cite{goodfellow2020generative} and variational autoencoders (VAEs) \cite{diederik2014auto}, have garnered significant attention for shape generation tasks.
Notably, the work by \cite{gao2019sdm} introduces a two-level variational autoencoder that simultaneously learns the overall shape structure and detailed part geometries. 
Both GRASS \cite{li2017grass} and StructureNet \cite{mo2019structurenet} employ techniques to compress the shape structure into a latent space while considering the relationships between different parts. 
Additionally, \cite{mo2020pt2pc} uses a part-tree decomposition to conditionally generate 3D shapes, and \cite{jones2020shapeassembly} adopts a procedural programmatic representation to establish connections between part cuboids. 
Inspired by the Seq2Seq networks in machine translation, \cite{wu2020pq} introduces a sequential encoding and decoding approach for the regression of shape parameters. 
While many of these approaches focus on directly generating new part shapes given random latent codes, our main focus revolves around the rigid transformation of existing parts to facilitate their assembly.


\section{Part-Whole-Hierarchy Message Passing}
\label{sec:model}
This section provides a detailed explanation of our proposed part-whole-hierarchy message passing network. 
We begin by presenting how we construct super-parts in an unsupervised manner.
Then we introduce our super-part encoder that predicts the latent poses of super-parts.
Subsequently, we describe how our part encoder leverages the latent poses and the part-whole hierarchy to predict part poses. 
The overall model is illustrated in Figure~\ref{fig:architecture}.
Finally, we introduce the loss functions and the training process for diverse generation.

\subsection{Super-Part Construction}

We denote the input point clouds as $\mathcal{P}=\{\mathbf{P}_i \vert i=1 \dots N \}$, where $\mathbf{P}_i \in \mathbb{R}^{d \times 3}$ corresponds to the point cloud of the $i$-th given part of the 3D shape.
The values of $N$ and $d$ represent the number of parts and the number of points per part respectively. 
Note that their values may differ from object to object. 
For notation convenience, we assume objects are padded to the maximum number of parts, and each part is padded to the maximum number of points per part.
The goal of our task is to predict a set of 6-DoF part poses $\left\{\left(t_i, r_i \right)  \right\}_{i=1}^{N}$, where $t_i \in \mathbb{R}^3$ and $r_i \in \mathbb{R}^4$ represent the translation and rigid rotation for each part, respectively. The complete part assembly for a 3D shape is $\mathcal{S}=\cup_{i=1}^{N} \mathbf{T}_i\left(\mathbf{P}_i  \right)$, and $\mathbf{T}_i$ represents a transformation in $SE(3)$ that consists of a 3D rotation in the rotation group $SO(3)$ and a 3D translation in the translation group, induced by $\left(t_i, r_i  \right)$.

A super-part is a subset of parts that is ideally semantically meaningful.
However, since we do not have ground-truth super-parts, we need to construct them in an unsupervised manner.
In particular, we compute axis-aligned bounding boxes for each part and evaluate the similarity between these 3D boxes. 
Parts are grouped into the same super-part if the difference between their respective enclosing boxes is below a specified threshold. Following previous works~\cite{zhan2020generative,zhang20223d}, we set the threshold to 0.2. 
Although the super-parts obtained through this method may lack semantic meaning, they provide a coarse abstraction grounded in geometry similarities without any supervised labels.
Then we group the input point cloud parts $\mathcal{P}=\left\{\mathbf{P}_i  \right\}_{i=1}^N$ into a set of $M$ super-parts $\mathcal{P}^{\prime}=\left\{\mathbf{P}^{\prime}_i  \right\}_{i=1}^M$ based on the part-whole hierarchy of a given object. 
Here $\mathbf{P}^{\prime}_i$ represents the $i$-th super-part.
To maintain consistent notation, we again pad the number of parts per super-part to the maximum $M$.
Note that padding is used here for clarity and understanding.
In practice, our model can handle sets of parts with varying sizes without padding.

\subsection{Super-Part Relation Reasoning}
To reason about the relationships among super-parts, we utilize the super-part encoder. 
The computational process is illustrated in the top half of Figure~\ref{fig:architecture}.

Given a set of point clouds of a super-part $\mathbf{P}^{\prime}_i$, we first compute the set-level permutation-invariant representation $\mathbf{F}^{\prime}_i$ via a shared PointNet \cite{qi2017pointnet}.
We use the same architecture and hyperparameters as in previous works~\cite{zhan2020generative,zhang20223d} for a fair comparison.
This feature representation captures the global characteristics of each super-part while being invariant to the order of the parts within the set.
Based on the features $\mathbf{F}^{\prime}_i$ extracted for each super-part from the PointNet, we use a two-layer multi-head attention (MHA) module
to learn the relationships between super-parts. 
Specifically, following \cite{vaswani2017attention}, we calculate an attention matrix $\mathbf{A'}$, where $\mathbf{A}^{\prime}_{i,j}$ represents the attention weight of the $i$-th super-part to the $j$-th super-part. Then we multiply each super-part feature $\mathbf{F}^{\prime}_j$ by $\mathbf{A}^{\prime}_{i,j}$ and sum them up to obtain the attention-weighted feature $\mathbf{G}^{\prime}_i$. 
In our model, we set the embedding dimension to 256 and the number of heads for the MHA module to 8.

For super-part pose prediction, we feed the attention-weighted feature $\mathbf{G}^{\prime}_i$ into a prediction head containing 4 fully-connected layers to obtain the pose $\left\{\left(t^{\prime}_i, r^{\prime}_i\right)\right\}_{i=1}^M$. 
We apply \emph{tanh} operation to the translation vector, which restricts the range of the part center offset as $(-1, 1)$. 
Additionally, for simplicity, we predict the quaternion vector $r^{\prime}_i=(r^{\prime}_{i0},r^{\prime}_{i1},r^{\prime}_{i2},r^{\prime}_{i3} )$ instead of the rotation matrix. 
We then use the $\textit{Rodrigues formula}$ \cite{rodrigues1840lois} to obtain the rotation matrix $\mathbf{R}^{\prime}_i$ corresponding to $r^{\prime}_i$ one by one as follows:
{\small
\begin{align}
    &\mathbf{R}^{\prime}_{i}= \nonumber 
    \begin{bmatrix}
    1-2{r^{\prime}_{i2}}^2-2{r^{\prime}_{i3}}^2,2r^{\prime}_{i1}r^{\prime}_{i2}-2r^{\prime}_{i0}r^{\prime}_{i3}, 2r^{\prime}_{i1}r^{\prime}_{i3}+2r^{\prime}_{i0}r^{\prime}_{i2} \\
    2r^{\prime}_{i1}r^{\prime}_{i2}+2r^{\prime}_{i0}r^{\prime}_{i3} ,1-2{r^{\prime}_{i1}}^2-2{r^{\prime}_{i3}}^2, 2r^{\prime}_{i2}r^{\prime}_{i3}-2r^{\prime}_{i0}r^{\prime}_{i1} \\
    2r^{\prime}_{i1}r^{\prime}_{i3}-2r^{\prime}_{i0}r^{\prime}_{i2} ,2r^{\prime}_{i2}r^{\prime}_{i3}+2r^{\prime}_{i0}r^{\prime}_{i1} ,1-2{r^{\prime}_{i1}}^2-2{r^{\prime}_{i2}}^2. \nonumber
    \end{bmatrix}
    \label{eq:quaternion}
\end{align}
}%
The transformed point cloud of the super-part is expressed as $\mathbf{T}_i\left(\mathbf{P}^{\prime}_i  \right)=\mathbf{R}^{\prime}_{i}\mathbf{P}^{\prime}_i + t^{\prime}_i$.
To ensure that it is a unit quaternion, we normalize the rigid rotation vector such that $\Vert r^{\prime}_i \Vert = 1$.



Since we do not have ground-truth super-part poses, the supervision for this module solely comes from the back-propagation signal from the subsequent module.
As a result, the super-part poses are considered as latent variables.

\subsection{Part Relation Reasoning}
As previously mentioned, the poses of super-parts offer valuable clues about the poses of their corresponding parts. 
To leverage this information, we design a part encoder to incorporate the latent super-part poses to predict the part poses.

\paragraph{Transformation via Latent Super-Part Poses}

Firstly, we apply the latent super-part poses to transform all point cloud parts, resulting in transformed parts denoted as $\hat{\mathcal{P}} = \{\mathbf{\hat{P}}_i\}_{i=1}^N$.
These transformed parts are then fed into the part encoder for further processing. 
Similarly to the super-part encoder, we employ another 
PointNet to extract part-level feature denoted as $\mathbf{\hat{F}}_i$. 
In order to better utilize the part-whole hierarchy of objects, we design two ways to integrate the super-part level information to the part level, \ie, cross-level and within-level attention modules.

\paragraph{Cross-Level Attention}
Part representations $\mathbf{\hat{F}}_i$ from the PointNet encoder does not leverage the part-whole hierarchy.
To better fuse the information from the super-parts, we propose a cross-level attention module.
Specifically, we treat the part representation $\mathbf{\hat{F}}_i$ as query and super-part representations $\mathbf{F}^{\prime}_i$ as keys and values. 
We then employ a MHA module to perform an attention-weighted aggregation of super-part representations $\mathbf{F}^{\prime}_i$ to obtain the updated part representations $\mathbf{F}_i$. 
By doing so, the information from part-whole hierarchy is explicitly integrated into part representations.
Here we specify the embedding dimension as 256 and the MHA module is set to utilize 8 heads.

\paragraph{Within-Level Attention}
To model the part relationships, we leverage another within-level attention module, which is depicted in the bottom half of Figure~\ref{fig:architecture}.  
We first concatenate the representation from the previous module $F_i$ with an additional instance encoding vector that is unique for each part following \cite{zhang20223d}.
This allows us to distinguish geometrically-equivalent parts. 
Then we perform message passing among parts using another MHA module where the attention computation is aware of the part-whole hierarchy.
Specifically, we concatenate the latent super-part poses to the part-level feature before computing the attention scores at each layer.
Consequently, in each message passing layer, a part receives pose hints from its parent super-part and updates its representation. 
The resulting part feature is denoted as $\mathbf{G}_i = \sum_{j=1}^N \mathbf{A}_{i,j} \mathbf{F}_i$, where $\mathbf{A}_{i,j}$ represents the attention weight of the $i$-th part to the $j$-th part.
This design enables the part-to-part message passing to effectively leverage the part-whole hierarchy information.
We set the number of heads for the MHA module to 8, the dimension of the part-level feature to 256, and concatenate it with the 40-dimensional instance encoding vector. 

\paragraph{Prediction Head}
Finally, we feed the attention-weighted part-level feature $\mathbf{G}_i$ to a part pose prediction head, which consists of four fully-connected layers. 
The channel sizes of the first three fully-connected layers are set to 256, 256 and 1024, respectively, followed by ReLU activation functions. 
We use a linear projection layer to predict the part pose $\left(t_i, r_i  \right)$, encompassing a 4-dimensional rigid rotation and 3-dimensional translation. 
This prediction head outputs the part poses $\left\{\left(t_i, r_i\right)\right\}_{i=1}^N$, where $t_i$ represents the translation vector and $r_i$ represents the rigid rotation vector for each part.
We adopt the same design as in the super-part encoder to restrict the translation vectors to the range of $(-1, 1)$ and normalize the rigid rotation vector $r_i$ to have a unit norm.

\subsection{Training Objective for Diverse Generation}
Considering the same set of input point cloud parts, geometrically equivalent parts (such as legs of a chair) can be interchanged and decorated parts may have multiple placement options, resulting in multiple possible shapes. 
For example, a semi-cylindrical part can be placed on top of a chair backrest as a headrest, or placed under the backrest as a lumbar pillow. 
To account for such diversified structural variations and configurations, we introduce random noise into the system following \cite{zhan2020generative,narayan2022rgl,zhang20223d} and employ the Min-of-N (MoN) loss \cite{fan2017point} to simultaneously consider assembly accuracy and assembly diversity. 
By considering the minimum of a set of N possible assemblies, we can encourage the model to generate diverse and valid shape configurations while maintaining overall accuracy in the assembly process. 
This helps address the challenge of capturing multiple plausible solutions that arise from the interchangeable and decoratable nature of the parts.


Let $\mathcal{F}\left(\mathcal{P},\delta_j \right)$ denote our network outputs and $\mathcal{F^{\ast}}\left(\mathcal{P}  \right)$ denote the ground-truth point clouds, then the MoN loss is:
\begin{equation}
    \mathcal{L}_{\text{MoN}}=\min\limits_{\delta_j\sim\mathcal{N}\left(0,1  \right)}\mathcal{L}\left(\mathcal{F}\left(\mathcal{P},\delta_j  \right), \mathcal{F^{\ast}}\left(\mathcal{P}  \right)   \right)
\end{equation}
where $\delta_j$ is a random noise vector drawn from standard Normal distribution in an IID fashion. 
Following \cite{zhang20223d}, we sample 5 random vectors $\delta_j$ during training. The loss function $\mathcal{L}$ consists of the following components for both local part and global shape losses.

Firstly, we supervise the translation via an $l_2$ loss between our prediction $t_i$ and the ground-truth translation vector $t_i^{\ast}$ for each part:
\begin{equation}
    \mathcal{L}_t = \sum\limits_{i=1}^N \|t_i-t_i^{\ast}    \|_2^2
\end{equation}

Then we compute the Chamfer distance \cite{fan2017point} between the predicted and the ground-truth rigid rotation for each part:
\begin{equation}
    \mathcal{L}_r = \sum\limits_{i=1}^N d_c\left(r_i\left(\mathbf{P}_i\right), r_i^{\ast}\left(\mathbf{P}_i  \right)   \right)
\end{equation}
The Chamfer distance $d_c\left(\mathcal{X},\mathcal{Y}\right)$ is a metric commonly used in point cloud comparison, which measures the dissimilarity between two point sets $\mathcal{X}$ and $\mathcal{Y}$ by calculating the average distance between each point in one set and its nearest neighbor in the other set as follows:
\begin{equation}
    d_c\left(\mathcal{X}, \mathcal{Y}   \right) = \sum\limits_{x\in \mathcal{X}}\min\limits_{y\in\mathcal{Y}}\|x-y   \|_2^2 + \sum\limits_{y\in \mathcal{Y}}\min\limits_{x\in\mathcal{X}}\|x-y   \|_2^2
    \label{eq:CD}
\end{equation}
In this case, we use the Chamfer distance to assess the discrepancy between the predicted rotation of each part and its corresponding ground-truth rotation. 

Similarly, the full shape $\mathcal{S}$, \ie, the set of point clouds belonging to all parts, is supervised via the Chamfer distance from the ground-truth shape $\mathcal{S}^{\ast}$:
\begin{equation}
    \mathcal{L}_s=d_c\left(\mathcal{S}, \mathcal{S}^{\ast}   \right)
    \label{eq:SCD}
\end{equation}

In summary, the overall loss function is defined as:
\begin{equation}
    \mathcal{L} = \lambda_t \mathcal{L}_t + \lambda_r \mathcal{L}_r + \lambda_s \mathcal{L}_s
    \label{eq:overall_loss}
\end{equation}
where $\lambda_t$, $\lambda_r$ and $\lambda_s$ represent the weights assigned to the three loss terms. 
Based on the cross validation, we empirically set $\lambda_t=1$, $\lambda_r=10$, and $\lambda_s=1$ in our experiments.
By minimizing this loss function, we aim to improve the alignment of the predicted poses with the ground-truth poses, thereby enhancing the overall accuracy of the assembly.

\section{Experiments}
\label{sec:experiments}
In this section, we demonstrate the effectiveness of the proposed model by comparing our results with state-of-the-art methods. We also provide visual analysis of the assembly process for super-parts and parts, which helps to illustrate the hierarchical assembly process from parts to the whole.

\begin{table*}
  \centering
  \resizebox{\textwidth}{!}{
  \begin{tabular}{lcccc|ccc|ccc|ccc|cccl}
    \toprule
    & & \multicolumn{3}{c}{SCD($10^{-2}$) $\downarrow$} & \multicolumn{3}{c}{PA(\%) $\uparrow$} & \multicolumn{3}{c}{CA(\%) $\uparrow$} & \multicolumn{3}{c}{QDS($10^{-5})$ $\uparrow$} & \multicolumn{3}{c}{WQDS($10^{-5})$ $\uparrow$}\\
    \midrule
    &Methods & Chair& Table& Lamp & Chair & Table & Lamp & Chair& Table& Lamp & Chair& Table& Lamp & Chair& Table& Lamp\\
    \midrule\midrule
    &B-Global \cite{li2020learning,schor2019componet} & 1.46     & 1.12    & 0.79 &  15.70  & 15.37 & 22.61 & 9.90  & 33.84  & 18.60 & 0.15 & 0.20 & 0.76 &1.25 & 1.40 &0.58\\
    &B-LSTM \cite{wu2020pq}& 1.31     & 1.25    & 0.77 &  21.77  & 28.64 & 20.78 & 6.80  & 22.56  & 14.05 & 0.04 & 0.27 & 0.63 &1.07 &1.43 &1.54\\
    &B-Complement \cite{sung2017complementme}& 2.41    &  2.98   & 1.50 & 8.78  & 2.32 & 12.67 & 9.19  & 15.57 & 26.56 & 0.09 & 0.06 & 2.81 &1.28 &1.75 &2.08\\
    &DGL \cite{zhan2020generative}& 0.91     & 0.50    & 0.93 &  39.00  & 49.51 & 33.33 & 23.87  & 39.96  & 41.70 & 1.69 & 3.05 & 1.84 &1.35 &2.97 & 1.73\\
    &Score \cite{cheng2023score}& 0.71     & 0.42    & 1.11 &  44.51  & 52.78 & 34.32 & 30.32  & 40.59  & 49.07 & 3.36 & 9.17 & 6.83 & 1.70 &3.81 & 2.82 \\
    &RGL \cite{narayan2022rgl}& 0.87  & 0.48    & 0.72 &  49.06  & 54.16 & 37.56 & 32.26  & 42.15  & 57.34 &3.55 & 7.63 &6.82 &2.12 &4.07 & 2.96 \\
    &IET \cite{zhang20223d}& 0.54    & 0.35    & 1.03 &  62.80  & 61.67 & 38.68 & 48.45  & 56.18  & 62.62 & 4.15 &9.09 & 6.98 & 2.74 &4.56 & 3.29 \\
    &Ours & \textbf{0.51} & \textbf{0.28}    & \textbf{0.70} &  \textbf{64.13}  & \textbf{64.83} & \textbf{38.80} & \textbf{49.28}  & \textbf{58.45}  & \textbf{64.16} & 
    \textbf{5.62} & 
    \textbf{9.58} & 
    \textbf{7.12} & 
    \textbf{3.06} & 
    \textbf{4.81} & 
    \textbf{3.90} 
    \\
    \bottomrule
  \end{tabular}}
  \vspace{-1mm}
  \caption{Comparison between our approach and other methods under the Chamfer distance threshold 0.01. 
  }
  \label{tab:q1}
  \vspace{-3mm}
\end{table*}

\subsection{Dataset}
Following~\cite{li2020learning,schor2019componet,wu2020pq,sung2017complementme,zhan2020generative,cheng2023score,narayan2022rgl,zhang20223d}, we use the PartNet \cite{mo2019partnet} dataset for both training and evaluation.
This dataset consists of 26,671 shapes across 24 different 3D object categories. 
To effectively validate and compare different methods, we select the three largest categories, \ie, chairs, tables and lamps, following~\cite{li2020learning,zhan2020generative,narayan2022rgl,zhang20223d,cheng2023score}.
In total, we have 6,323 chairs, 8,218 tables, 2,207 lamps in the finest-grained level, and we adopt the official train/validation/test splits (70\%/10\%/20\%) to conduct the experiments. 
For each part point cloud, 1000 points are sampled from the original part meshes using Farthest Point Sampling (FPS) \cite{moenning2003fast}.
All parts are transformed into its canonical space using PCA.

\subsection{Evaluation Metrics}
We generate a variety of shapes by adding different Gaussian noises to a given set of input parts, and find the closest shape to the ground-truth using minimum matching distance (MMD)~\cite{achlioptas2018learning}. 
To measure part assembly quality, we follow \cite{zhan2020generative,narayan2022rgl,zhang20223d} to use \textit{shape Chamfer distance} (SCD), \textit{part accuracy} (PA) and \textit{connectivity accuracy} (CA). 
\textit{Shape Chamfer distance} is defined in Equation \eqref{eq:CD} and Equation \eqref{eq:SCD}. 
Meanwhile, to compare the diversity of assembled parts, we propose two quantitative evaluation metrics including quality-diversity score (QDS) and the weighted quality-diversity score (WQDS) following~\cite{cheng2023score}.
We also visualize the generation results to qualitatively evaluate the diversity.
More details of these metrics are provided in the appendix.



\paragraph{Mean Part/Connectivity Accuracy (mPA/mCA).}
PA and CA depend on the Chamfer distance threshold $\tau_{p}$ and $\tau_{c}$ for judging whether the assembly is accurate. 
In order to provide a more comprehensive evaluation of the performance of the model, we average the results under multiple thresholds to get \textit{mean part accuracy} (mPA) and \textit{mean connectivity accuracy} (mCA), formally, 
\begin{equation}
    \text{mPA} = \frac{1}{T_{p}} \sum\limits_{\tau_{p}\in T_{p}}\text{PA}\left(\tau_p \right), \quad
    \text{mCA} = \frac{1}{T_{c}} \sum\limits_{\tau_{c}\in T_{c}}\text{CA}\left(\tau_c \right), \nonumber
\end{equation}
where $T_{p}$ and $T_{c}$ are set to $\left\{0.01, 0.02, 0.03, 0.04, 0.05  \right\}$. 


\paragraph{QDS.} 
Diversity score (DS)~\cite{mo2020pt2pc,shu20193d} evaluates the diversity of the results as $\text{DS}=\frac{1}{N^2}\sum\nolimits_{i,j=1}^N(d_c(\mathbf{P}_i^{\ast}, \mathbf{P}_j^{\ast}))$, where $\mathbf{P}_i^{\ast}$ and $\mathbf{P}_j^{\ast}$ represent any two assembled shapes. 
Based on DS, we propose the quality-diversity score (QDS) as below,
{\small
\begin{align}
    \text{QDS}=\frac{1}{N^2}\sum\limits_{i,j=1}^N d_c(\mathbf{P}_i^{\ast}, \mathbf{P}_j^{\ast}) \mathbbm{1} (\text{CA}(\mathbf{P}_i^{\ast}>\tau_q))  \mathbbm{1} (\text{CA}(\mathbf{P}_j^{\ast}>\tau_q)). \nonumber
\end{align}}%
QDS imposes constraints to remove pairs that are of low assembly quality, thus assessing not only diversity but also the quality of generated shapes. 
Following \cite{cheng2023score}, we adopt $\text{SCD}$ as the distance metric. 
The value of $\tau_q$ is set to 0.5 in both QDS and WQDS.

\paragraph{WQDS.} Based on QDS, we further propose the weighted quality-diversity score (WQDS) as below,
{\small
\begin{align}
    \text{WQDS}=\frac{1}{N^2}\sum\limits_{i,j=1}^N & d_c(\mathbf{P}_i^{\ast}, \mathbf{P}_j^{\ast}) \text{CA}(\mathbf{P}_i) \text{CA}(\mathbf{P}_j) \nonumber  \\
        & \mathbbm{1} (\text{CA}(\mathbf{P}_i^{\ast}>\tau_q)) \mathbbm{1} (\text{CA}(\mathbf{P}_j^{\ast}>\tau_q)). \nonumber
    \label{eq:WQDS}
\end{align}}%
WQDS weights the QDS of a pair by their connectivity accuracy, thus favoring assembled shapes that demonstrate a high-quality connection between each pair of parts. 

\subsection{Comparisons with State-of-the-Art}

\begin{figure*}
  \centering
  \includegraphics[width=1.0\textwidth]{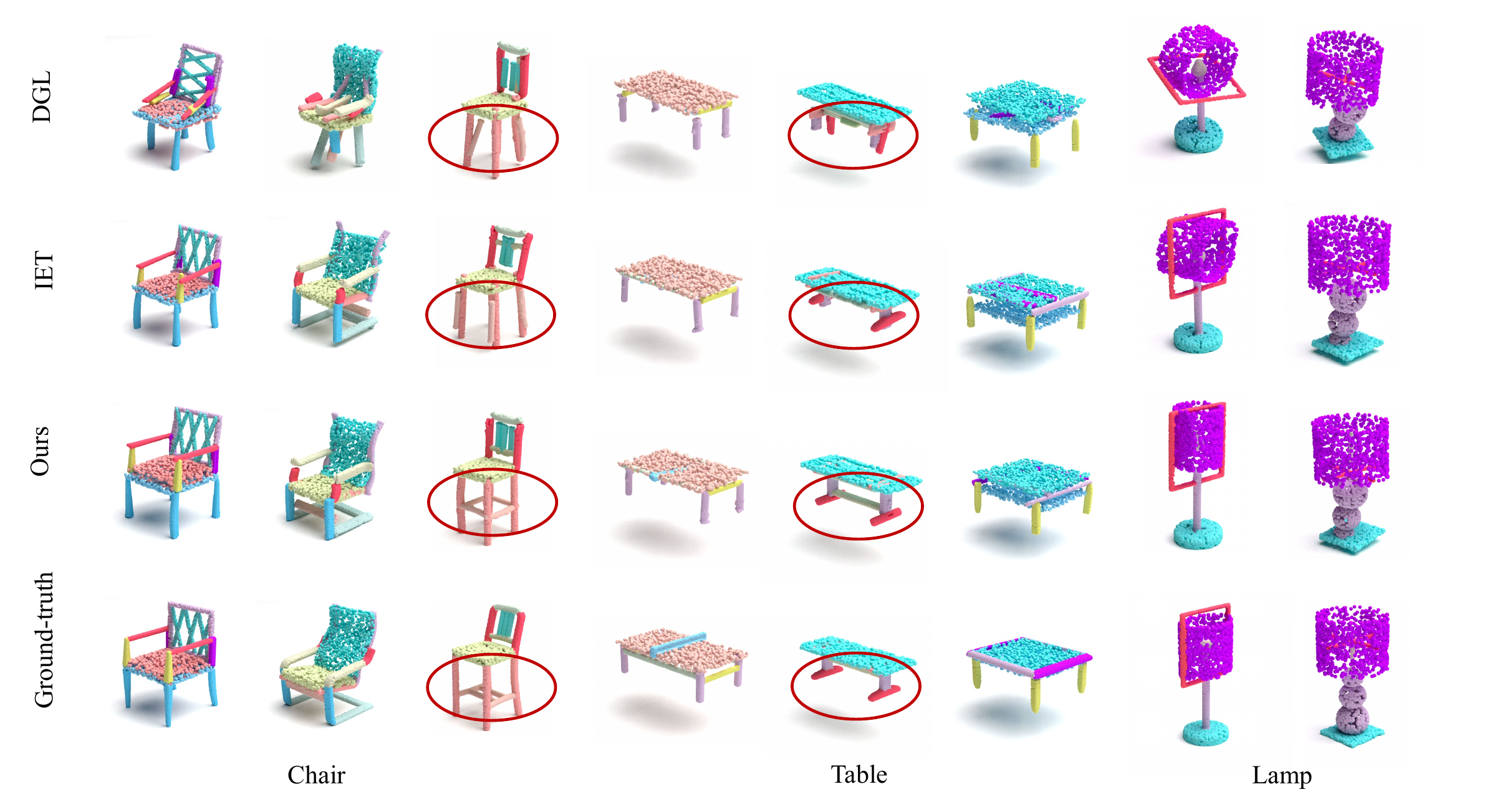}
  \vspace{-8mm}
  \caption{The qualitative comparisons between our method and two most competitive baselines on PartNet \cite{mo2019partnet}. We highlight some areas where the assembly quality of ours is clearly better. 
  }
  \label{fig:mainres}
  \vspace{-5mm}
\end{figure*}


\begin{table}
  \centering
  \resizebox{\columnwidth}{!}{
  \begin{tabular}{lc|ccc|cccl}
    \toprule
    & \multicolumn{1}{c}{} & \multicolumn{3}{c}{mPA $\uparrow$} & \multicolumn{3}{c}{mCA $\uparrow$} \\
    \midrule
    &Methods & Chair & Table & Lamp & Chair& Table& Lamp \\
    \midrule\midrule
    &DGL \cite{zhan2020generative} & 45.84  & 57.86 & 48.32  & 29.17   & 43.88   & 51.75\\
    &IET \cite{zhang20223d} & 74.92  & 73.20  & 52.80 & 62.37  & 68.49 & 75.94  \\
    &Ours  &  \textbf{76.79}  & \textbf{76.19}  & \textbf{53.31} & \textbf{65.32} &  \textbf{70.26} &  \textbf{78.15} \\
    \bottomrule
  \end{tabular}}
  \vspace{-1mm}
  \caption{Comparison between our approach and other methods under multiple Chamfer distance thresholds from 0.01 to 0.05. 
  }
  \label{tab:q2}
  \vspace{-3mm}
\end{table}

\begin{figure}
  \centering
  \includegraphics[width=1.0\columnwidth]{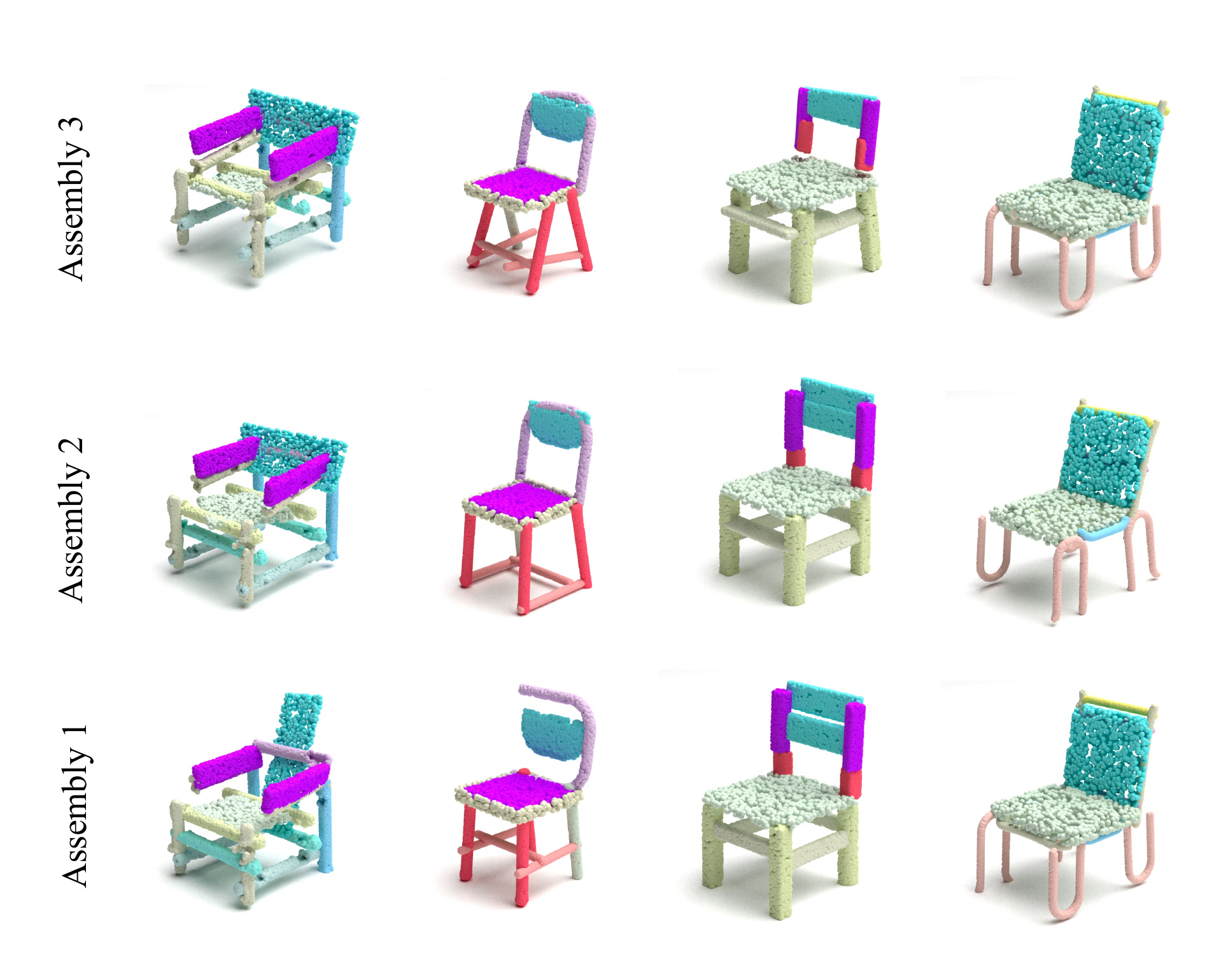}
  \vspace{-8mm}
  \caption{Diverse results on the unseen PartNet\cite{mo2019partnet} test dataset generated by our network to demonstrate the structural variation in part assembly, providing different artistic results while maintaining reasonable object structures.}
  \label{fig:diversity}
  \vspace{-5mm}
\end{figure}

We compare our method with the state-of-the-art methods on PartNet \cite{mo2019partnet} dataset. 
Following previous works, we first present the quantitative results under a certain Chamfer distance threshold of 0.01 in Table \ref{tab:q1}.
As we can see, our method consistently outperforms all competitors.
We also show the visualization of assembly results in Figure \ref{fig:mainres}.
It is clear that our method is better than others in terms of both the coherence of the assembled structure and the connectivity between parts, \eg, the positioning of crossbeams between chair legs, and the connection between table legs and the tabletop.

\begin{figure*}
  \centering
  \includegraphics[width=1.0\textwidth]
  {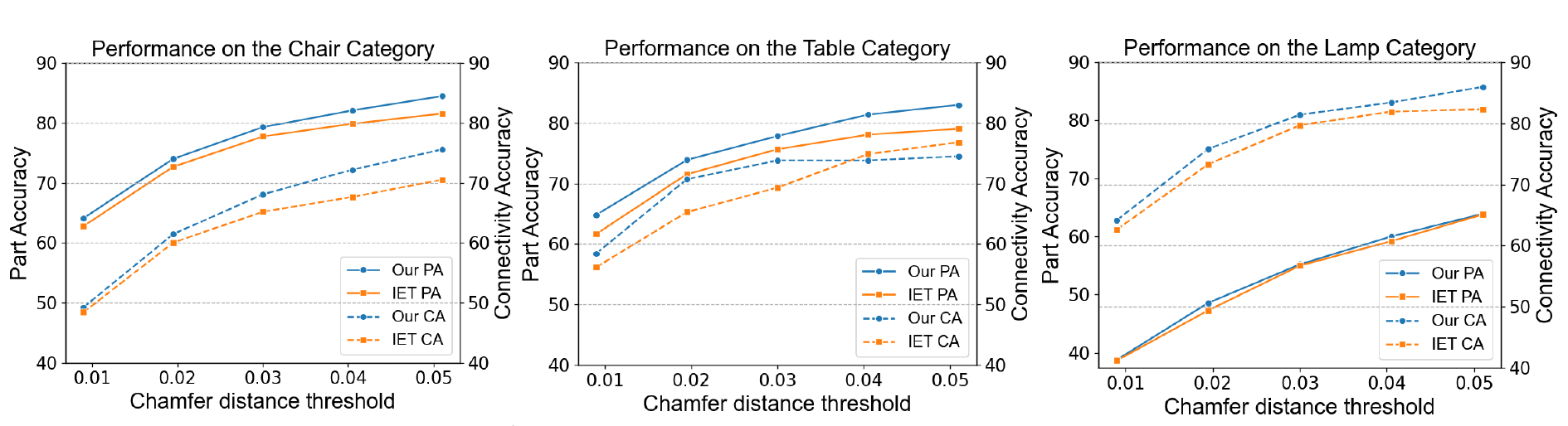}
  \vspace{-5mm}
  \caption{Performance on the \textbf{Chair}, \textbf{Table} and \textbf{Lamp} categories under multiple Chamfer distance thresholds.}
  \label{fig:threshold}
  \vspace{-3mm}
\end{figure*}

To eliminate the sensitivity of performances with respect to threshold values and provide a more comprehensive evaluation, we compare the mPA and mCA of our method with two competitive baselines in Table \ref{tab:q2}.
For the \textbf{Chair} category, our method demonstrates a significant improvement over IET \cite{zhang20223d}, with a 1.87\% increase in mPA and a 2.95\% increase in mCA, while the improvement in the \textbf{Lamp} category is relatively minor. 
This may be attributed to the limited variation in the geometry of lamp parts, making it difficult to clearly differentiate between various levels of the part-whole hierarchy.
In addition, Figure \ref{fig:threshold} illustrates the performance of various methods on the \textbf{Chair} and \textbf{Table} categories using five different Chamfer distance thresholds ranging from 0.01 to 0.05. As the Chamfer distance threshold increases, the advantage of our method becomes more pronounced, which shows the superiority of our approach.
We also evaluate the diversity of generative part assembly. 
As shown in Table~\ref{tab:q1}, our method outperforms all competing methods in both QDS and WQDS. 
In Figure~\ref{fig:diversity}, we use the model to generate multiple assembly shapes for the same set of parts (with different noises) to show the variation of generated shapes.

\subsection{Human Study}
Since many inaccurate results are still valid visually, we also conduct a human study to evaluate the quality of generated assemblies. Specifically, we perform an A/B test with a group of students, utilizing the identical evaluation system as in~\cite{xiong2023learning,zyrianov2022learning}. Participants were presented with pairs of randomly chosen assemblies of the same object from two different methods. In total, 28 participants labeled 93 assembly pairs. We compare the proposed method with three main baselines and present the outcomes in Table~\ref{tab:humanstudy}. In over 95\% of the cases, our method is favored over other three competing methods, unequivocally demonstrating the superior visual quality of our method's output.

\subsection{Ablation Study}
We now investigate the importance of different loss components and verify the effectiveness of the proposed super-part encoder on the Chair category.

\begin{table}  
  \centering
  \resizebox{0.8\columnwidth}{!}{%
  \begin{tabular}{lc|ccc}
    \toprule
    & Settings &  SCD$\downarrow$ & PA$\uparrow$ & CA$\uparrow$ \\
    \midrule
    &  \begin{tabular}{@{}c@{}}w/o Super-Part Enc. \\ + Augmented Param.\end{tabular} & 0.0038 & 60.70  & 53.46  \\
    \midrule
    & w/o Super-Part Enc. & 0.0036 & 61.04 & 55.39 \\
    \midrule
    & Full Setting & \textbf{0.0028}  & \textbf{64.83} & \textbf{58.45} \\
    \bottomrule
  \end{tabular}
  }
  \vspace{-1mm}
  \caption{Ablation studies on part-whole hierarchy on the $\textbf{Table}$ category under a certain Chamfer distance threshold 0.01.}
  \label{tab:ablationhierarchy}  
\end{table}

\begin{table}
  \centering
  \resizebox{0.8\columnwidth}{!}{%
  \begin{tabular}{lccc|ccc}
    \toprule
    & $\mathcal{L}_t$ & $\mathcal{L}_r$ & $\mathcal{L}_s$ &  SCD$\downarrow$ & PA$\uparrow$ & CA$\uparrow$ \\
    \midrule
    &  & $\checkmark$ & $\checkmark$ &  0.0075 & 35.36 & 32.73 \\
    \midrule
    & $\checkmark$ &  & $\checkmark$ &  0.0039 & 60.02 & 54.21 \\
    \midrule
    & $\checkmark$ & $\checkmark$ &  &  0.0033 & 62.74 & 56.98 \\
    \midrule
    & $\checkmark$ & $\checkmark$ & $\checkmark$ & \textbf{0.0028} & \textbf{64.83} & \textbf{58.45} \\
    \bottomrule
  \end{tabular}
  }
  \vspace{-1mm}
  \caption{Ablation studies on loss components on the $\textbf{Table}$ category under a certain Chamfer distance threshold 0.01.}
  \label{tab:ablationloss}  
\end{table}

\begin{table}  
  \centering  
  \resizebox{0.8\columnwidth}{!}{%
  \begin{tabular}{lc|c}
    \toprule
    & Method &  Percent Prefer Ours \\
    \midrule
    &  Ours vs IET\cite{zhang20223d} &  95.2\% \\
    \midrule
    &  Ours vs RGL \cite{narayan2022rgl} &  97.6\% \\
    \midrule
    &  Ours vs DGL \cite{zhan2020generative} & 99.5\% \\
    \bottomrule
  \end{tabular}
  }
  \vspace{-1mm}
  \caption{Human study results on PartNet\cite{mo2019partnet}.}
  \label{tab:humanstudy}  
\end{table}

\begin{figure}
  \centering
  \includegraphics[width=1.0\columnwidth]{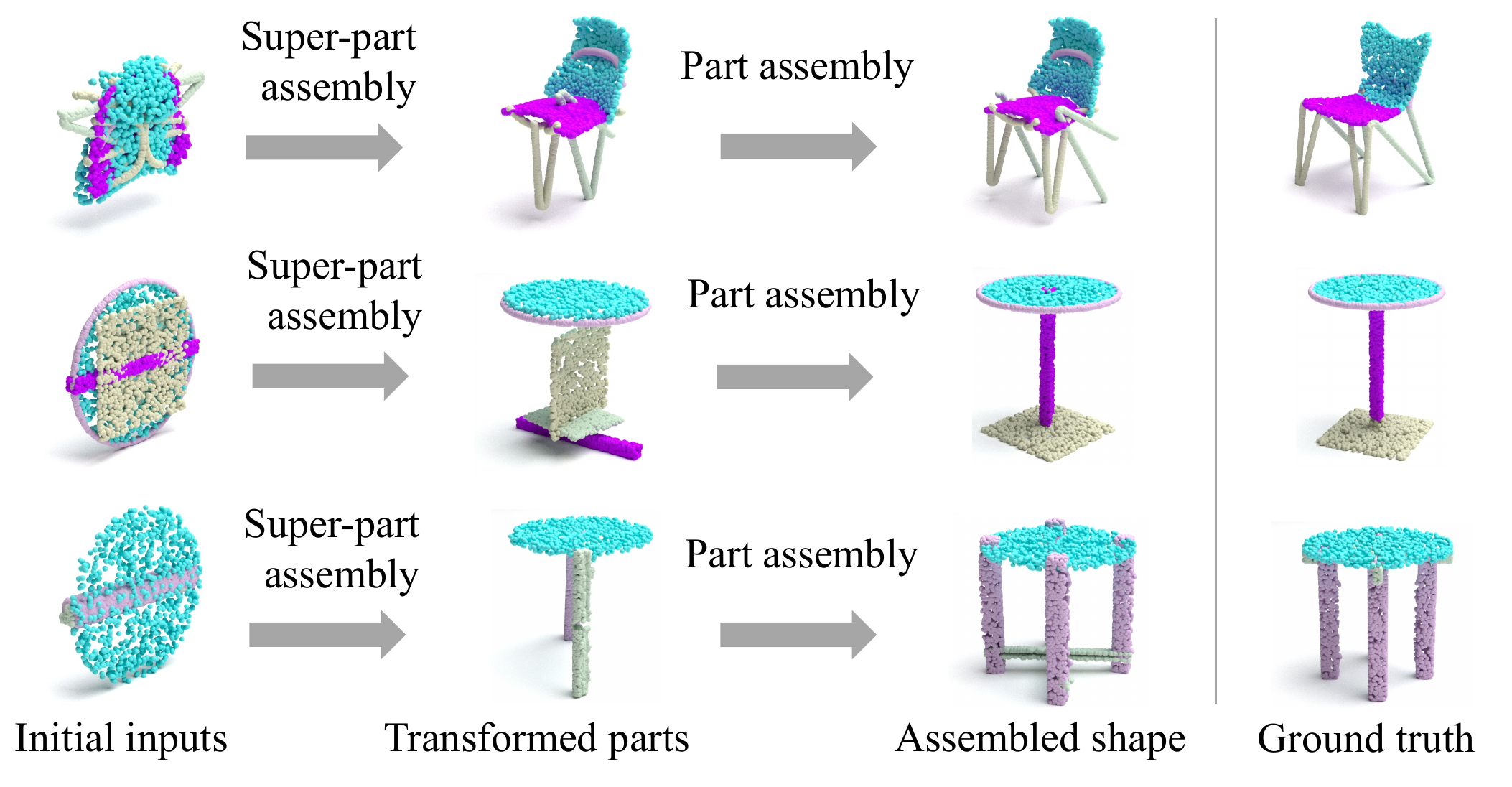}
  \vspace{-5mm}
  \caption{Visualization of our predicted hierarchical assembly process from parts to the whole. 
  }
  \label{fig:trajectory}
  \vspace{-5mm}
\end{figure}

\paragraph{Super-Part Encoder.}
Since the super-part encoder is at the core of our part-whole hierarchy message passing network, we conduct an experiment where we remove this module (denoted as \textbf{w/o Super-Part Enc.}).
To mitigate the impact of the number of model parameters, we also create another baseline by increasing the number of parameters of the \textbf{w/o Super-Part Enc.} setting (denoted as \textbf{w/o Super-Part Enc. + Augmented Param.}).
This involves increasing the number of parameters in the part encoder and the multi-head attention module, ensuring that the overall number of parameters is on the same order of magnitude as the full model setting.
As shown in Table \ref{tab:ablationhierarchy}, the super-part encoder contributes significantly to the final performance.
We also observe that augmenting the model parameters without incorporating the super-part encoders actually leads to a performance drop.
This finding further highlights the effectiveness of the proposed part-whole hierarchy network for part assembly.


\paragraph{Loss Functions.}
We now evaluate the importance of individual loss terms in Equation \eqref{eq:overall_loss}.
As shown in Table \ref{tab:ablationloss}, removing the part translation loss $\mathcal{L}_t$ leads to a significant performance degradation.
Similarly, the remaining loss terms, including the rigid rotation loss $\mathcal{L}_r$ and the shape assembly loss $\mathcal{L}_s$, demonstrate their significance in facilitating precise part rotations and ensuring the overall coherence of the assembled parts, respectively.
This shows that each loss term does play an indispensable role in achieving high-quality assemblies.


\subsection{Hierarchical Part Assembly Analysis}
We further provide a visual analysis of the assembly process for super-parts and parts in Figure \ref{fig:trajectory}. 
It is obvious that during the super-part assembly phase, important components such as chair seats and chair backs are often assembled correctly first. 
This aligns with our intuition that super-part assembly is relative easier and provides strong hints for predicting poses of parts like chair legs and arms, thus showing further evidence for the effectiveness of the part-whole-hierarchy message passing network.

\section{Conclusion}\label{sec:conclusion}
In this paper, we propose the part-whole-hierarchy message passing network to address the challenging generative 3D part assembly task.
We first group the point cloud of individual parts to form super-parts in an unsupervised way.
Taking the point cloud as input, our super-part encoder predicts latent poses of super-parts which are used to transform the point cloud.
We then feed the transformed point cloud to the part encoder.
Relying on the cross-level and within-level attention based message passing, the part encoder takes the transformed point cloud as input and leverages the information from super-parts and predicts poses of parts.
Experiments on the PartNet dataset demonstrate that our method achieves state-of-the-art performances as well as provides interpretable assembly process. 
In the future, we are interested in combining 3D part generation with our 3D part assembly model to generate 3D shapes from scratch.

\subsubsection*{Acknowledgments}
This work was funded, in part, by NSERC DG Grants (No. RGPIN-2022-04636), the Vector Institute for AI, and Canada CIFAR AI Chair. 
Resources used in preparing this research were provided, in part, by the Province of Ontario, the Government of Canada through the Digital Research Alliance of Canada \url{alliance.can.ca}, and companies sponsoring the Vector Institute \url{www.vectorinstitute.ai/#partners}, and Advanced Research Computing at the University of British Columbia. 
Additional hardware support was provided by John R. Evans Leaders Fund CFI grant.

\clearpage

{
    \small
    \bibliographystyle{ieeenat_fullname}
    \bibliography{main}
}


\end{document}